\documentclass[runningheads,a4paper]{llncs}

\usepackage{amssymb}
\usepackage{graphicx}
\usepackage{ifpdf}
\usepackage{amsfonts}
\usepackage{graphicx}
\usepackage{epsfig}
\usepackage{mathtools}
\usepackage{bm}
\usepackage{mathrsfs}
\usepackage{color}
\usepackage{multirow}
\usepackage[linesnumbered,ruled]{algorithm2e}

\usepackage{url}
\urldef{\mailsa}\path|{alfred.hofmann, ursula.barth, ingrid.haas, frank.holzwarth,|
\urldef{\mailsb}\path|anna.kramer, leonie.kunz, christine.reiss, nicole.sator,|
\urldef{\mailsc}\path|erika.siebert-cole, peter.strasser, lncs}@springer.com|

\begin{document}

\mainmatter  

\title{Training Multi-organ Segmentation Networks with Sample Selection by Relaxed Upper Confident Bound}

\titlerunning{RUCB}

\author{Yan Wang$^1$, Yuyin Zhou$^{1}$, Peng Tang$^{2}$, \\
Wei Shen$^{1,3}$, Elliot K. Fishman$^{4}$, Alan L. Yuille$^1$}
\authorrunning{Y. Wang \emph{et al.}}
\institute{$^1$Johns Hopkins University $^2$Huazhong University of Science and Technology $^3$Shanghai University $^4$The Johns Hopkins University School of Medicine}


%
%


%
%

\maketitle

\vspace{-3mm}
\begin{abstract}
Deep convolutional neural networks (CNNs), especially fully convolutional networks, have been widely applied to automatic medical image segmentation problems, e.g., multi-organ segmentation. Existing CNN-based segmentation methods mainly focus on looking for increasingly powerful network architectures, but pay less attention to data sampling strategies for training networks more effectively. In this paper, we present a simple but effective sample selection method for training multi-organ segmentation networks. Sample selection exhibits an exploitation-exploration strategy, i.e., exploiting hard samples and exploring less frequently visited samples. Based on the fact that very hard samples might have annotation errors, we propose a new sample selection policy, named Relaxed Upper Confident Bound (RUCB). Compared with other sample selection policies, e.g., Upper Confident Bound (UCB), it exploits a range of hard samples rather than being stuck with a small set of very hard ones, which mitigates the influence of annotation errors during training. We apply this new sample selection policy to training a multi-organ segmentation network on a dataset containing 120 abdominal CT scans and show that it boosts segmentation performance significantly.
\end{abstract}
\vspace{-5mm}

\section{Introduction}
The field of medical image segmentation has made significant advances riding on the wave of deep convolutional neural networks (CNNs). Training convolutional deep networks (CNNs), especially fully convolutional networks (FCNs)~\cite{Ref:Long15}, to automatically segment organs from medical images, such as CT scans, has become the dominant method, due to its outstanding segmentation performance. which also sheds lights to many clinical applications, such as diabetes inspection, organic cancer diagnosis, and surgical planning.

To approach human expert performance, existing CNN-based segmentation methods mainly focus on looking for increasingly powerful network architectures, e.g., from plain networks to residual networks~\cite{Ref:FakhryZJ17,Ref:RonnebergerFB15}, from single stage networks to cascaded networks~\cite{Ref:ShenWJWY17}, from networks with a single output to networks with multiple side outputs~\cite{Ref:MerkowMKT16,Ref:ShenWJWY17}. However, there is much less study of how to select training samples from a fixed dataset to boost performance.

\begin{figure}[t]
\centering
\includegraphics[width=0.9\textwidth]{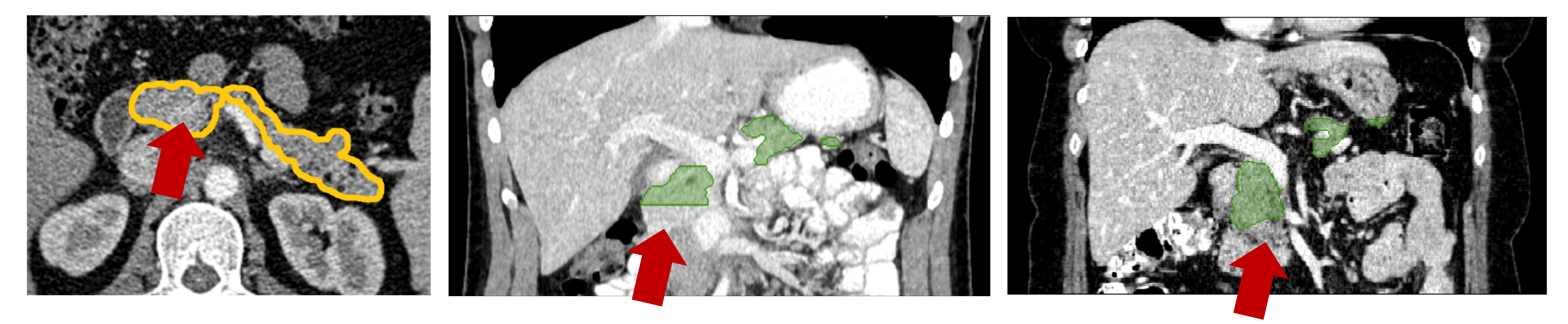}
\caption{Some examples in a abdominal CT scans dataset which have annotations errors. Left: vein is included in pancreas segmentation; Middle $\&$ Right: missing pancreas header\vspace{-1em}.}
\label{fig:errorsample}
\end{figure}

In the training procedure of current state-of-the-art CNN-based segmentation method~\cite{Ref:ZhouXSWFY17,Ref:RothLFSS16,Roth2017Hierarchical3F,Ref:Guo2018,Ref:CicekALBR16}, training samples (2D slices for 2D FCNs and 3D sub-volumes for 3D FCNs) are randomly selected to iteratively update network parameters. However, some samples are much more difficult to segment than others, e.g., those which contain more organs with indistinct boundaries or with very small sizes. It is known that using hard sample selection, or called bootstrapping{\footnote{In this paper, we only consider the bootstrapping procedure that selects samples from a fixed dataset.}}, for training deep networks yields faster training, higher accuracy, or both~\cite{Ref:ShrivastavaGG16,Ref:LoshchilovH15,Ref:Simo-SerraTFKM14}. Hard sample selection strategies for object detection~\cite{Ref:ShrivastavaGG16} and classification~\cite{Ref:LoshchilovH15,Ref:Simo-SerraTFKM14} base their selection on the training loss for each sample, but some samples are hard may due to annotation errors, as shown in Fig.~\ref{fig:errorsample}. This problem may not be significant for the tasks in natural images, but for the tasks in medical images, such as multi-organ segmentation, usually requires very high accuracy, and thus the annotation errors are more significant. Our experiments show that the training losses of samples (such as the samples in Fig.~\ref{fig:errorsample}) with annotation errors are very large, and even larger than real hard samples.

To address this problem, we propose a new hard sample selection policy, named Relaxed Upper Confident Bound (RUCB). Upper Confident Bound (UCB)~\cite{Ref:AuerCF02} is a classic policy to deal with \emph{exploitation}-\emph{exploration} trade-offs~\cite{Ref:Auer02}, e.g., exploiting hard samples and exploring less frequently visited samples for sample selection. UCB was used for object detection in natural images~\cite{Ref:CanevetF16}, but UCB is easy to be stuck with some samples with very large losses, as the selection procedure goes on. In our RUCB, we relax this policy by selecting hard samples from a larger range, but with higher probability for harder samples, rather than only selecting some very hard samples as the selection procedure goes on. RUCB can escape from being stuck with a small set of very hard samples, which can mitigate the influence of annotation errors. Experimental results on a dataset containing 120 abdominal CT scans show that the proposed Relaxed Upper Confident Bound policy boosts multi-organ segmentation performance significantly.

\vspace{-1mm}
\section{Methodology}

Given a 3D CT scan $V =(v_j, j=1,...,|V|)$, the goal of multi-organ segmentation is to predict the label of all voxels in the CT scan $\hat{{Y}}=(\hat{y}_j, j = 1,...,|V|)$, where $\hat{y}_j \in \{0, 1, ..., |\mathcal{L}|\}$ denotes the predicted label for each voxel $v_j$, i.e., if $v_j$ is predicted as a background voxel, then $\hat{y}_j=0$; and if $v_j$ is predicted as an organ in the organ space $\mathcal{L}$, then $\hat{y}_j = 1, ..., |\mathcal{L}|$. In this section, we first review the basics of the Upper Confident Bound policy~\cite{Ref:AuerCF02}, then elaborate our proposed Relaxed Upper Confident Bound policy on sample selection for multi-organ segmentation.
\vspace{-5mm}
\subsection{Upper Confident Bound (UCB)}
\label{sec:ucb}
The Upper Confident Bound (UCB)~\cite{Ref:AuerCF02} policy is widely used to deal with the exploration versus exploitation
dilemma, which arises in the multi-armed bandit (MAB) problem \cite{Ref:Robbins52}. In a $K$-armed bandit problem, each arm $k=1,...,K$ is recorded by an unknown distribution associated with an unknown expectation. In each trial $t=1,...,T$, the learner $\mathcal{A}$ takes an action to choose one of $K$ alternatives $g(t)\in\{1,...,K\}$  and collects a reward $x_{g(t)}^{(t)}$. The objective of this problem is to maximize the long-run cumulative expected reward $\sum_{t=1}^Tx_{g(t)}^{(t)}$. But, as the expectations are unknown, the learner can only make a judgement based on the record of the past trails.

At trial $t$, the UCB selects the alternative $k$ maximizing
\begin{equation}
\bar{x}_k + \sqrt{\frac{2\ln n}{n_k}},
\end{equation}
where $\bar{x}_k={\sum_{t=1}^{n} x_k^{(t)}}/{n_k}$ is the average reward obtained from the alternative $k$ based on the previous trails, $x_k^{(t)}=0$ if $x_k$ is not chosen in the $t$-th trail. $n_k$ is the number of times alternative $k$ has been selected so far and $n$ is the total number of trail done. The first term is the exploitation term, whose value is higher if the expected reward is larger; and the second term is the exploration term, which grows with the total number of actions have been taken but shrinks with the number of times this particular action have been tried. At the beginning of the process, the exploration term dominates the selection, but as the selection procedure goes on, the one with the best expected reward will be chosen.

\vspace{-3mm}
\subsection{Relaxed Upper Confident Bound (RUCB) Boostrapping}
Fully convolutional networks (FCNs)~\cite{Ref:Long15} are the most popular model for multi-organ segmentation. In a typical training procedure of an FCN, a sample (e.g., a 2D slice) is randomly selected in each iteration to calculate the model error and update model parameters. To train this FCN more effectively, a better strategy is to use hard sample selection, rather than random sample selection. As sample selection exhibits an exploitation-exploration trade-off, i.e., exploiting hard samples and exploring less frequently visited samples, we can directly apply UCB to select samples, where the reward of a sample is defined as the network loss function w.r.t. it. However, as the selection procedure goes on, only a small set of samples with the very large reward will be selected for next iteration according to UCB. The selected sample may not be a proper hard sample, but a sample with annotation errors, which inevitably exist in medical image data as well as other image data. Next, we introduce our Relaxed Upper Confident Bound (RUCB) policy to address this issue\vspace{-1em}.

\subsubsection{Procedure}

We consider that training an FCN for multi-organ segmentation, where the input images are 2D slices from axial directions. Given a training set $\mathcal{S}=\{(\mathbf{I}_i,\mathbf{Y}_i)\}_{i=1}^M$, where $\mathbf{I}_i$ and $\mathbf{Y}_i$ denote a 2D slice and its corresponding label map, and $M$ is the number of the 2D slices, like the MAB problem, each slice $\mathbf{I}_i$ is set to be associated with the number of times it was selected $n_i$ and the average reward obtained through the training $\bar{J}_i$.
After training an initial FCN with randomly sampling slices from the training set, FCN is boostrapped several times by sampling hard and less frequently visited slices. In the sample selection procedure, rewards are assigned to each training slice once, then the next slice on which to train FCN is chosen by the proposed RUCB. The reward of this slice returned by FCN is then fed into RUCB and updates the statistics in RUCB. This process is then repeated to select another slice based on the new updated statistics, until a max-iteration $N$ is reached. Statistics are reset to 0 before beginning a new boostrapping phase because slices that are chosen in previous rounds may no longer be informative\vspace{-1em}.

\subsubsection{Relaxed Upper Confident Bound}

We denote the corresponding label map of the input 2D slice $\mathbf{I}_i\subset\mathbb{R}^{H\times W}$ as $\mathbf{Y}_i=\{y_{i,j}\}_{j=1,...,H\times W}$. If $\mathbf{I}_i$ is selected to update the FCN in the $t$-th iteration, the reward obtained for $\mathbf{I}_i$ is computed by
\begin{equation}
\label{eq:score}
\mathcal{J}^{(t)}_i(\mathbf{\Theta})=-\frac{1}{H\times W}\left [ \sum_{j=1}^{H\times W}\sum_{l=0}^{|\mathcal{L}|}\mathbf{1}\left ( y_{i,j}=l \right )\log p^{(t)}_{i,j,l} \right ],
\end{equation}
where $p_{i,j,l}^{(t)}$ is the probability that the label of the $j$-th pixel in the input slice is $l$, and $p_{i,j,l}^{(t)}$ is parameterized by the network parameter $\mathbf{\Theta}$.
If $\mathbf{I}_i$ is not selected to update the FCN in the $t$-th iteration, $\mathcal{J}^{(t)}_i(\mathbf\Theta)=0$. After $n$ iterations, the next slice to be selected by UCB is the one maximizing $\bar{J}_i^{(n)}+\sqrt{{2\ln n}/{n_i}}$, where $\bar{J}_i^{(n)}=\sum_{t=1}^{n}\mathcal{J}^{(t)}_i(\mathbf\Theta)/{n_i}$.

{\renewcommand\baselinestretch{0.4}\selectfont
\begin{algorithm}[t]
\label{alg:RUCB}
    \SetKwInOut{Input}{Input}
    \SetKwInOut{Output}{Output}
    \Input{FCN parameter $\mathbf\Theta$, input training slices $\{\mathbf{I}_i\}_{i=1,...,M}$\;
    \qquad\qquad parameters $\alpha$ and $\beta$\;
     \qquad\qquad max number of iterations $T$;}
    \Output{FCN parameter $\mathbf\Theta$;}
    total number of times slices are selected $n\leftarrow 0$\;
    number of times slice $\mathbf{I}_1,...,\mathbf{I}_m$ are selected $n_1, ...., n_m\leftarrow 0$\;
    running index $i\leftarrow 0$\;
    $\mathcal{J}_1^{(1)}, ..., \mathcal{J}_M^{(M)}\leftarrow 0$\;
    \Repeat{$n=M$}
    {$i\leftarrow i+1$, $n_i\leftarrow n_i + 1$, $n\leftarrow n+1$\;
      Compute $\mathcal{J}_i^{(i)}$ by Eq.~\ref{eq:score}\;
      $\bar{J}_i^{(M)}=\sum_{t=1}^M\mathcal{J}^{(i)}_i/n_i$\;}
   $\forall i$, compute $\tilde{J}_i^{(M)}$ by Eq.~\ref{eq:normalization}, compute $q_i^{(M)}$ by Eq.~\ref{eq:ucbscore2}\;
   $\mu=\sum_{i=1}^M{q}_i^{(M)} /M$, $\sigma=\text{std}({q}_i^{(M)})$\;

   iteration $t\leftarrow 0$\;
   \Repeat{$t=T$}
   {
    $t\leftarrow t+1$\;
$\alpha\sim \mathcal{U}(0,a)$\;
$K=\sum_{i=1}^M(\mathbf{1}({q}_i^{(M)}>\mu + \alpha\sigma))$\;
randomly select a slice $\mathbf{I}_i$ from the set $\{\mathbf{I}_i|q_i^{(n)}\in\mathcal{D}_K(\{q_i^{(n)}\}_{i=1}^M)\}$\;
$n_i\leftarrow n_i+1$, $n\leftarrow n+1$\;
Compute $\mathcal{J}_i^{(t)}$ by Eq.~\ref{eq:score}, $\mathbf\Theta\leftarrow\arg\min_{\mathbf\Theta} {\mathcal{J}^{(t)}_i(\mathbf\Theta)}$\;
$\bar{J}_i^{(n)}=\sum_{t=1}^n\mathcal{J}^{(t)}_i/n_i$\;
$\forall i$, compute $\tilde{J}_i $ by Eq.~\ref{eq:normalization}, compute $q_i^{(n)}$ by Eq.~\ref{eq:ucbscore2}\;
   }

\caption{Relaxed Upper Confident Bound}
\end{algorithm}
\par}

Preliminary experiments show that reward defined above is usually around $[0,0.35]$. The exploration term dominates the exploitation term. We thus normalize the reward to make a balance between exploitation and exploration by

\begin{equation}
\label{eq:normalization}
\tilde{J}_i^{(n)}=\min\left\{\beta, \frac{\beta}{2}\frac{\bar{J}_i^{(n)}}{\sum_{i=1}^{{M}} \bar{J}_i^{(n)}/{M}} \right \},
\end{equation}
where the $\min$ operation ensures that the score lies in $[0, \beta]$. Then the UCB score for $\mathbf{I}_i$ is calculated as
\begin{equation}
\label{eq:ucbscore2}
q_i^{(n)} = \tilde{J}_i^{(n)}+\sqrt{\frac{2\ln n}{n_i}}.
\end{equation}

As the selection procedure goes on, the exploitation term of Eq.~\ref{eq:ucbscore2} will dominate the selection, i.e., only some very hard samples will be selected. But, these hard samples may have annotation errors. In order to alleviate the influence of annotation errors, we propose to introduce more randomness in UCB scores to relax the largest loss policy. After training an initial FCN with randomly sampling slices from the training set, we assign an initial UCB score $q_i^{(M)}=\tilde{J}_i^{(M)}+\sqrt{2\ln M/1}$ to each slice $\mathbf{I}_i$ in the training set. Let us assume the UCB scores of all samples follow a normal distribution $\mathcal{N}(\mu, \sigma)$. Hard samples are regarded as slices whose initial UCB scores are larger than $\mu$. Note that initial UCB scores are only decided by the exploitation term. In each iteration of our bootstrapping procedure, we count the number of samples that lie in the range $[\mu+\alpha\cdot\text{std}({q}_i^{(M)}),+\infty]$, denoted by $K$, where $\alpha$ is drawn from a uniform distribution $[0, a]$ ($a=3$ in our experiment), then a sample is selected randomly from the set $\{\mathbf{I}_i|q_i^{(n)}\in\mathcal{D}_K(\{q_i^{(n)}\}_{i=1}^M)\}$ to update the FCN, where $\mathcal{D}_K(\cdot)$ denote the K largest values in a set. Here we count the number of hard samples according to a dynamic range, because we do not know the exact range of hard samples. This dynamic region enables our bootstrapping to select hard samples from a lager range with higher probability for harder samples, rather than only selecting some very hard samples. We name our sample selection policy Relaxed Upper Confident Bound (RUCB), as we choose hard samples in a larger range, which introduces more variance to the hard samples. The training procedure for RUCB is summarized in Algorithm~\ref{alg:RUCB}.

\vspace{-1mm}
\section{Experimental Results}
\vspace{-1mm}
\subsection{Experimental Setup}
\vspace{-1mm}
\subsubsection{Dataset:} We evaluated our algorithm on 120 abdominal CT scans of normal cases under IRB (Institutional Review Board) approved protocol. CT scans are contrast enhanced images in portal venous phase, obtained by Siemens SOMATOM Sensation64 and Definition CT scanners, composed of (319-1051) slices of $(512 \times 512)$ images, and have voxel spatial resolution of $([$0.523-0.977$] \times [$0.523-0.977$] \times $0.5$)mm^{3}$. Sixteen organs (including aorta, celiac AA, colon, duodenum, gallbladder, interior vena cava, left kidney, right kidney, liver, pancreas, superior mesenteric artery, small bowel, spleen, stomach, and large veins) were segmented by four full-time radiologists, and confirmed by an expert. This dataset is a high quality dataset, but a small portion of error is inevitable, as shown in Fig.~\ref{fig:errorsample}. Following the standard corss-validation strategy, we randomly partition the dataset into four complementary folds, each of which contains 30 CT scans. All experiments are conducted by four-fold cross-validation, i.e., training the models on three folds and testing them on the remaining one, until four rounds of cross-validation are performed using different partitions\vspace{-1em}.

\subsubsection{Evaluation Metric:} The performance of multi-organ segmentation is evaluated in terms of Dice-S{\o}rensen similarity coefficient (DSC) over the whole CT scan. We report the average DSC score together with the standard deviation over all testing cases\vspace{-1em}.

\subsubsection{Implementation Details: } We use FCN-8s model \cite{Ref:Long15} pre-trained on PascalVOC in caffe toolbox. The learning rate is fixed to be 1$\times$$10^{-9}$ and all the networks are trained for 80$K$ iterations. Three boostrapping phases are conducted, at 20,000, 40,000 and 60,000 respectively, i.e., the max number of iterations for each boostrapping phase is $T=20,000$. We set $\beta=2$, since $\sqrt{2\ln n/n_i}$ is in the range of [3.0, 5.0] in boostrapping phases\vspace{-1em}.

\subsection{Evaluation of RUCB}

We evaluate the performance of the proposed sampling algorithm (RUCB) with other competitors. Three sampling strategies considered for comparisons are (1) uniform sampling (Uniform); (2) online hard example mining (OHEM) \cite{Ref:ShrivastavaGG16}; and (3) using UCB policy (i.e., select the slice with the largest UCB score during each iteration) in boostrapping.

\begin{table}[t]
\centering
\label{tbl:results}
\scriptsize
\setlength{\tabcolsep}{2mm}
\renewcommand\arraystretch{1}
\begin{tabular}{l|c |c| c| c}
\hline
Organs     & Uniform &  OHEM &  UCB &  RUCB (ours) \\
\hline
Aorta			&    81.53 $\pm$ 4.50 & 77.49 $\pm$ 5.90             &  81.02 $\pm$ 4.50    & 81.03 $\pm$ 4.40\\
Adrenal gland		&    29.33 $\pm$ 16.26 & 31.44 $\pm$ 16.71             &  33.75 $\pm$ 16.26    & 36.76 $\pm$ 17.28\\
Celiac AA			&    34.49 $\pm$ 12.92 & 33.34 $\pm$ 13.86            &  35.89 $\pm$ 12.92    & 38.45 $\pm$ 12.53\\
Colon			&    77.51 $\pm$ 7.89 & 73.20 $\pm$ 8.94             &  76.40 $\pm$ 7.89    & 77.56 $\pm$ 8.65\\
Duodenum		&    63.39 $\pm$ 12.62 & 59.68 $\pm$ 12.32             &  63.10 $\pm$ 12.62    & 64.86 $\pm$ 12.18\\
Gallbladder		&    79.43 $\pm$ 23.77 & 77.82 $\pm$ 23.58             &  79.10 $\pm$ 23.77   & 79.68 $\pm$ 23.46\\
IVC				&    78.75 $\pm$ 6.54 & 73.73 $\pm$ 8.59             &  77.10 $\pm$ 6.54   & 78.57 $\pm$ 6.69\\
Left kidney		&    95.35 $\pm$ 2.53 & 94.24 $\pm$ 8.95             &  95.53 $\pm$ 2.53   & 95.57 $\pm$ 2.29\\
Right kidney		&    94.48 $\pm$ 9.49 & 94.23 $\pm$ 9.19             &  94.39 $\pm$ 9.49    & 95.40 $\pm$ 3.62\\
Liver				&    96.03 $\pm$ 1.70 & 90.43 $\pm$ 4.74             &  95.68 $\pm$ 1.70    & 96.00 $\pm$ 1.28\\
Pancreas			&    77.86 $\pm$ 9.92 & 75.32 $\pm$ 10.42             &  78.25 $\pm$ 9.92    & 78.48 $\pm$ 9.86\\
SMA				&    45.36 $\pm$ 14.36 & 47.18 $\pm$ 12.75           &  44.63 $\pm$ 14.36    & 49.59 $\pm$ 13.62\\
Small bowel		&    72.35 $\pm$ 13.30 & 67.44 $\pm$ 13.22             &  72.16 $\pm$ 13.30    & 72.88 $\pm$ 13.98\\
Spleen			&    95.32 $\pm$ 2.17 & 94.56 $\pm$ 2.41             &  95.16 $\pm$ 2.17    & 95.09 $\pm$ 2.44\\
Stomach			&    90.62 $\pm$ 6.51 & 86.37 $\pm$ 8.53             &  90.70 $\pm$ 6.51    & 90.92 $\pm$ 5.62\\
Veins			&    64.95 $\pm$ 19.96 & 60.87 $\pm$ 19.02             &  62.70 $\pm$ 19.96    & 65.13 $\pm$ 20.15\\
\hline
AVG			&    73.55 $\pm$ 10.28   &   71.08 $\pm$ 11.20   &   73.47 $\pm$ 10.52   &   74.75 $\pm$ 9.88   \\
\hline
\end{tabular}
\caption{DSC (\%) of sixteen segmented organs (mean $\pm$ standard deviation)\vspace{-1em}.}
\end{table}

Table~\ref{tbl:results} summarizes the results for 16 organs. Experiments show that images with wrong annotations are with large rewards, even larger than real hard samples after training an initial FCN. The proposed RUCB outperforms over all baseline algorithms in terms of average DSC. We see that RUCB achieves much better performance for organs such as \emph{Adrenal gland} (from 29.33$\%$ to 36.76$\%$), \emph{Celiac AA} (34.49$\%$ to 38.45$\%$), \emph{Duodenum} (63.39$\%$ to 64.86$\%$), \emph{Right kidney} (94.48$\%$ to 95.40$\%$), \emph{Pancreas} (77.86$\%$ to 78.48$\%$) and \emph{SMA} (45.36$\%$ to 49.59$\%$), compared with Uniform, which also shows our dataset is in high quality, where only a small portion of errors are contained. Most of the organs listed above are small organs which are difficult to segment. Besides, RUCB achieves 74.07$\pm$9.84 after 60$K$ iterations, which is even better than UCB trained in 80$K$ iterations. This also shows the efficiency of RUCB.

OHEM performs worse than Uniform, suggesting that directly sampling among slices with largest average rewards during boostrapping phase cannot help to train a better FCN. UCB obtains even slightly worse DSC compared with Uniform, as it only focuses on some hard examples which may have errors.

\begin{figure}[t]
\centering
\includegraphics[width=0.75\textwidth]{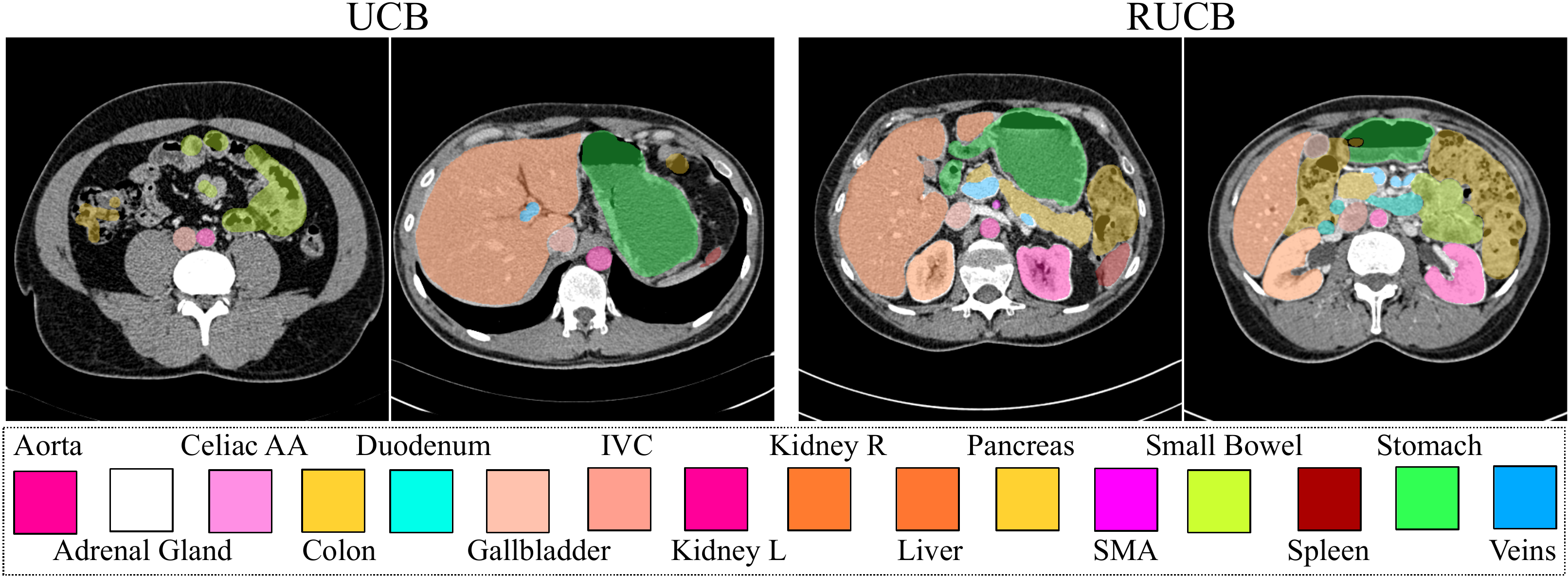}
\caption{Visualization of samples selected frequently by left: UCB and right: RUCB. Ground-truth annotations are marked in different colors\vspace{-2em}.}
\label{fig:visualize}
\end{figure}

To better understand UCB and RUCB, some of the hard samples selected more frequently are shown in Fig.~\ref{fig:visualize}. Some slices selected by UCB contain obvious errors such as \emph{Colon} annotation for the first one. Slices selected by RUCB are very hard to segment since it contains many organs including very small ones\vspace{-1em}.

\vspace{-3mm}
\subsubsection{Parameter Analysis} $\alpha$ is an important hyper-parameter for our RUCB. We vary it in the following range: $\alpha\in\{0,1,2,3\}$, to see how the performance of some organs changes. The DSC of \emph{Adrenal gland} and \emph{Celiac AA} are 35.36$\pm$17.49 and 38.07$\pm$12.75, 32.27$\pm$16.25 and 36.97$\pm$12.92, 34.42$\pm$17.17 and 36.68$\pm$13.73, 32.65$\pm$17.26 and 37.09$\pm$12.15, respectively. Using a fixed $\alpha$, the performance decreases. We also test the results when $K$ is a constant number, i.e., $K=5000$. The DSC of \emph{Adrenal gland} and \emph{Celiac AA} are 33.55$\pm$17.02 and 36.80$\pm$12.91. Compared with UCB, the results further verify that relaxing the UCB score can boost the performance\vspace{-1mm}.

\section{Conclusion}
\vspace{-1mm}
We proposed Relaxed Upper Confident Bound policy for sample selection in training multi-organ segmentation networks, in which the exploitation-exploration trade-off is reflected on one hand by the necessity for trying all samples to train a basic classifier, and on the other hand by the demand of assembling hard samples to improve the classifier. It exploits a
range of hard samples rather than being stuck with a small set of very hard samples, which mitigates the influence of annotation errors during training. Experimental results showed the effectiveness of the proposed RUCB sample selection policy.
\vspace{-2mm}
\bibliographystyle{splncs03}
\bibliography{mybibfile}
\end{document}